\documentclass[10pt,conference,a4paper]{IEEEtran}

\usepackage{url}

\usepackage{amsmath}
\usepackage{amssymb}

\usepackage{bm}

\usepackage{graphicx}
\usepackage{caption}
\usepackage{subcaption}
\graphicspath{{./images/}}

\usepackage[usenames, dvipsnames]{color}
\usepackage{stackengine}
\usepackage{multirow}
\usepackage{balance}

\usepackage[colorlinks=true,urlcolor=blue,linkcolor=black,citecolor=blue,bookmarks=true,pdfauthor={Johanna Carvajal, Arnold Wiliem, Conrad Sanderson, Brian Lovell},pdftitle={Towards Miss Universe Automatic Prediction: The Evening Gown Competition}]{hyperref}

\def\Vec#1{{\boldsymbol{#1}}}
\def\Mat#1{{\boldsymbol{#1}}}

\begin{document}

\title{\Large\bf Towards Miss Universe Automatic Prediction: The Evening Gown Competition}

\author{\IEEEauthorblockN{
		Johanna Carvajal{\tiny ~}$^{\dagger\diamond}$, 
		Arnold Wiliem{\tiny ~}$^{\dagger}$, 
		Conrad Sanderson{\tiny ~}$^{\dagger\diamond}$,		
		Brian Lovell{\tiny ~}$^{\dagger}$}
\IEEEauthorblockA{
{\tiny ~}$^{\dagger}$~University of Queensland, Brisbane, Australia\\
{\tiny ~}$^{\diamond}$~Data61, CSIRO, Australia\\
}
}

\maketitle

\begin{abstract}
Can we predict the winner of Miss Universe after watching how they stride down the catwalk during the evening gown competition? 
Fashion gurus say they can! 
In our work, we study this question from the perspective of computer vision. 
In particular, we want to understand whether existing computer vision approaches can be used to automatically extract the qualities exhibited by the Miss Universe winners during their catwalk.
This study can pave the way towards new vision-based applications for the fashion industry.
To this end, we propose a novel video dataset, called the Miss Universe dataset, comprising 10 years of the evening gown competition selected between 1996-2010.
We further propose two ranking-related problems: (1) Miss Universe Listwise Ranking and (2) Miss Universe Pairwise Ranking.
In addition, we also develop an approach that simultaneously addresses the two proposed problems. 
To describe the videos we employ the recently proposed Stacked Fisher Vectors in conjunction with robust local spatio-temporal features.
From our evaluation we found that although the addressed problems are extremely challenging, the proposed system is able to rank the winner in the top 3 best predicted scores for 5 out of 10 Miss Universe competitions.

\end{abstract}

\section{Introduction}

Miss Universe is a  worldwide pageant competition held every year since 1952 and is organised by \textit{The Miss Universe Organization}~\cite{miss_universe}. 
Every year up to 89 candidates participate in the competition. Each delegate must first win  their respective national pageants.
Miss Universe is broadcast  in more than 190 countries around the world and is watched by more than half a billion people annually~\cite{miss_universe,mu_facts2}. 
The format has slightly changed during the 64 year period. However, the most common competition format is as follows. All candidates are preliminary judged in three areas of competition: Interview, Swimsuits and Evening Gown. After that, the top 10 or 15 semi-finalists are short-listed during the coronation night. The semi-finalists compete again in swimsuits and evening gowns. The best 5 finalists are selected and go through an interview round. Finally, the runners-up and winner are announced. 

Although Miss Universe is one of the most publicised beauty pageants in the world, it is not the only existing pageant competition. 
A list of beauty pageants from around the world includes up to 22 events among international, continental and, regional pageants. 
Moreover, there are more than 260 national pageants. 
In the US alone, there are approximately 28 national pageants~\cite{list_beauty_pageants}.

During the swimsuit and evening gown competition, the catwalk is judged by  several aspects. 
Candidates must emanate poise, posture, grace, elegance, balance, confidence, energy, charisma, and sophistication. 
Additionally, during the swimsuit competition candidates are expected to have a well-proportioned body, good muscle tone, proper level of body fat and show fitness and body shape. 
In our work, we aim to capture these qualities  to predict the winner. 
This can  pave the way of numerous vision-based applications for the fashion industry such as automatic training systems for amateur models who aspire to become professionals.
Due to the complexity of this problem, we propose to initially study the evening gown competition. 
To this end, we collect a new dataset of videos recorded during the evening gown 
competition where the judges' scores are publicly available.

As mentioned, there are many potential commercial application for an automatic system able to analyse and predict the best catwalk in a beauty pageant. Automatically predicting the winner can be useful 
for specialised betting sites such as Odds Shark, Sports Bet, Bovada, and Bet Online. 
These betting sites allow the audience to bet for their favourite candidate in Miss Universe.  
Moreover, the catwalk analysis can be a powerful tool for boutique talent agencies such as {\it Polished by Donna} that 
provides training for improving the catwalk and offer their services to future beauty pageants candidates~\cite{polishedbydonna}.
For boutique talent agencies, an automatic catwalk analysis system
can help to compare the catwalk of each amateur model against herself or against an experienced catwalker.

Towards automatic prediction of Miss Universe, we first collect a novel Miss Universe (MU) dataset. 
The dataset comprises 10 years of Miss Universe selected from 1996 to 2010.
Only those years with available videos and official scores are selected.
The years included in this datasets are: 1996, 1997, 1998, 1999, 2000, 2001, 2002, 2003, 2007, and 2010.
The years not included were due to the videos and/or the scores were not publicly available.

It comprises 105 videos and 18,343 frames depicting each candidate catwalk in the evening gown competition. 
Fig.~\ref{fig:cat_walk} shows two examples of best and worst judges' scores during the evening gown competition.
We propose two sub-problems: (1) Miss Universe Listwise Ranking (MULR), and (2) Miss Universe Pairwise Ranking (MUPR).
The MULR problem aims to predict the winner of the evening gown competition, which can be useful during a beauty pageant competition and also for betting sites.
The MUPR problem focuses on judging the catwalk between two participants or to see the improvement of one model's catwalk.
The solution of the MUPR problem could be used for developing applications for boutique talent agencies.

In this work, we propose an approach which will address both problems simultaneously.
More specifically, we found that it is possible to share the model trained from one problem with the other problem.
We use our approach in conjunction with the video descriptors used for action analysis in~\cite{Joha2014b}.
In particular the video descriptors are extracted on a pixel-base and make use of gradients and optical flow. Gradients and optical flow have been shown to be effective for video representation.
Then, the video descriptors are encoded using the Stacked Fisher Vectors (SFV) approach, which has recently shown successful performance for action analysis~\cite{Peng2014eccv}.
From our evaluations, we found that that our proposed problems are extremely challenging. 
However, further analysis suggests that both problems could still be potentially solved 
using a computer vision approach.

\textbf{Contributions --- } we present 4 main contributions: {\bf (1)}~we study two novel problems for automatic ranking of Miss Universe evening gown competition participants using computer vision techniques; {\bf (2)}~we propose a novel dataset called the Miss Universe (MU) dataset that comprises 10 years of the Miss Universe evening gown competition selected between 1996-2010; {\bf (3)}~we propose an approach that addresses both problems simultaneously; {\bf (4)}~we adapt recent video descriptors, shown to be effective in action analysis, into our framework.

We continue our paper as follows. Section~\ref{sec:related_work} summarises the related work. The two sub-problems for  automatically predicting the winner of Miss Universe during the evening gown competition are explained in Section~\ref{sec:problem_def}.
In Section~\ref{sec:proposed_method}  we present our proposed approach that simultaneously addresses the two problems.
Section~\ref{sec:dataset} describes the Miss Universe dataset, the evaluation protocol, and the evaluation metrics.
In Section~\ref{sec:experiments} we present the results for both sub-problems. The main findings are summarised in Section~\ref{sec:conl}.

\section{Related Work}\label{sec:related_work}

An automatic system to predict the best catwalk in a beauty pageant has not been investigated before. The catwalk during the evening gown competition can be seen as a walk assessment, action assessment, or fine-grained action analysis. For catwalk assessment in Miss Universe, judges assess the quality of the walk. Catwalk analysis can be also related to fine-grained action analysis, where the aim is to distinguish the fine and subtle differences between two candidate catwalks. In the following sections, we summarise recent works for catwalk Assessment, Fine-Grained Action Analysis and some popular features for action analysis.

\textbf{Action Assessment --- } Gait and walk assessments have been investigated for elderly people and humans with neurological disorders~\cite{FangWang2013,Gholami2015}. 
Two web-cams are used to extract gait parameters including walking speed, step time, and step length in~\cite{FangWang2013}. 
The gait parameters are used for a fall risk assessment tool for  home monitoring of older adults.
For rehabilitation and treatment of patients with neurological disorders, automatic  gait analysis with a Microsoft Kinect sensor is used to quantify the gait abnormality of
patients with multiple sclerosis~\cite{Gholami2015}.
A gait analysis system consisting of two camcoders located on the right and left side of a treadmill is employed in~\cite{Nguyen2014}. This system fully reconstructs the skeleton model and demonstrates good accuracy compared to Kinect sensors.
Despite  being a related problem, for our Miss Universe catwalk analysis, Kinect sensors or   multi-cameras are simply not available.
The assessment of quality of actions using only visual information is still under early development. A recent work to predict the expert judges' scores for actions diving and figure skating in the Olympic games is presented in~\cite{Pirsiavash2014}. The concept behind the score prediction is to learn how to assess the quality of actions in videos.

\textbf{Fine-Grained Action Analysis --- } Catwalk analysis can be also related to fine-grained action analysis. Fine-grained action analysis has been recently investigated for action recognition~\cite{Joha2016a,Cheron2015,Kataoka2015,Pishchulin2014,Rohrbach2012}, where it is important to recognise small differences in activities such as cut and peel in food preparation. This is in contrast  to traditional action recognition where the goal is to recognise full-body activities such as walking or jumping.

\textbf{Features for Action Analysis --- } Improved dense trajectory (IDT) features in conjunction with Fisher Vector representation have recently show outstanding performance for the action recognition problem~\cite{HengWang2013}. 
This approach densely samples feature points at several spatial scales in each frame and tracks them using optical flow.  For each trajectory the following descriptors are computed: Trajectory, Histogram of Gradients, Histogram of Optical Flow, and Motion Boundary histogram. Finally, all descriptors are concatenated and normalised. IDT features are also popular for fine-grained action recognition~\cite{Kataoka2015, Pishchulin2014, Rohrbach2012}. However, some disadvantages have been reported. IDT generates irrelevant trajectories that are eventually are discarded. Processing such trajectories is time consuming and hence not suitable for realistic environments~\cite{Haiam2015,Hao2013}. 

Gradients have been used as a relatively simple yet effective video representation~\cite{Joha2014b}. 
Each pixel in the gradient image helps extract relevant information, eg.~edges of a subject. Gradients can be computed at every spatio-temporal location {\small $(x, y, t)$} in any direction in a video. 
Lastly, since the task of action recognition is based on an ordered sequence of frames, optical flow can be used to provide an efficient way of capturing local dynamics and motion patterns in a scene~\cite{Kliper2012}.

\section{Problem Definition}\label{sec:problem_def}

\begin{figure*}[tb!]
\centering
\begin{minipage}[c]{0.8\textwidth}
 \begin{minipage}[c]{0.15\textwidth}
  ~
\centering{\small Best catwalk}
\\\vspace{15mm}
\centering{\small Worst catwalk}

 \end{minipage}
 \begin{minipage}[c]{0.4\textwidth}
   \centering{\small Miss Universe 2003}
   \\\vspace{1mm}
   \includegraphics[width=0.32\columnwidth]{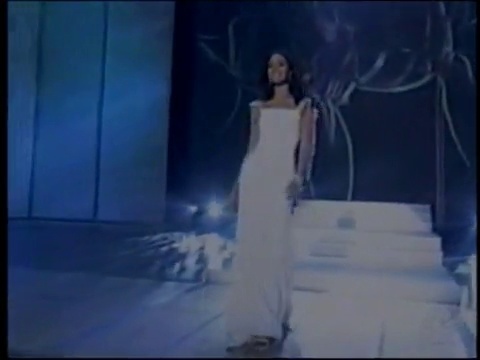}  
   \includegraphics[width=0.32\columnwidth]{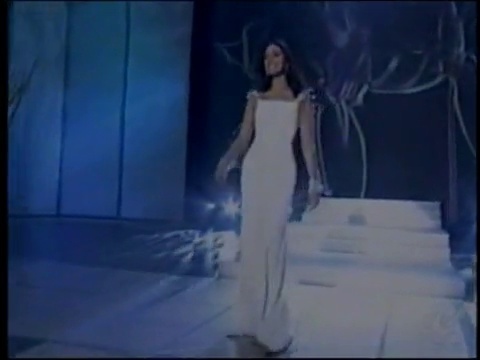}  
   \includegraphics[width=0.32\columnwidth]{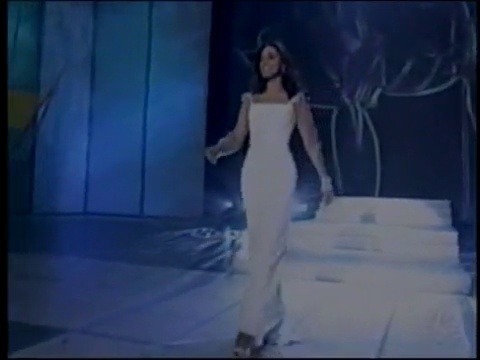}  
   \\\vspace{1mm}
  \includegraphics[width=0.32\columnwidth]{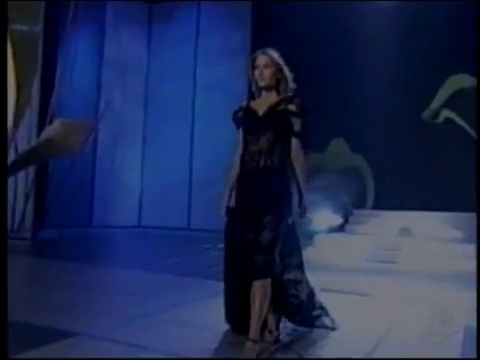}  
  \includegraphics[width=0.32\columnwidth]{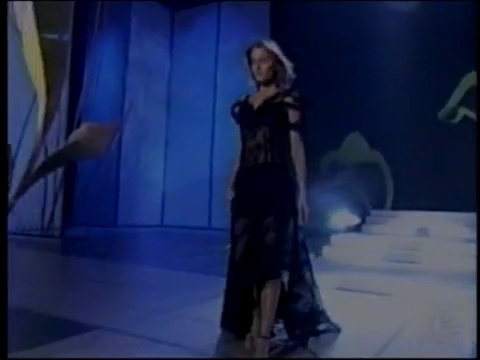}  
  \includegraphics[width=0.32\columnwidth]{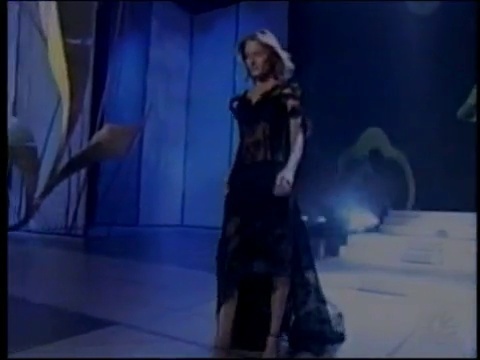} 
 \end{minipage}
 ~\vrule~
  \begin{minipage}[c]{0.4\textwidth}
  \centering{\small Miss Universe 2010}
  \\\vspace{1mm}
   \includegraphics[width=0.32\columnwidth]{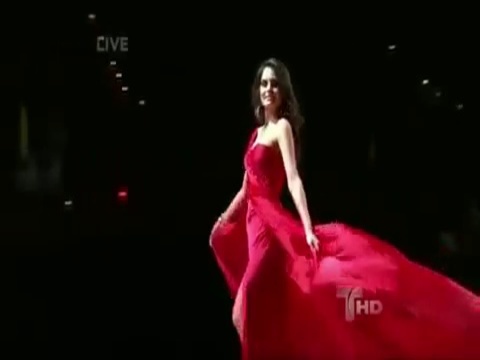}
   \includegraphics[width=0.32\columnwidth]{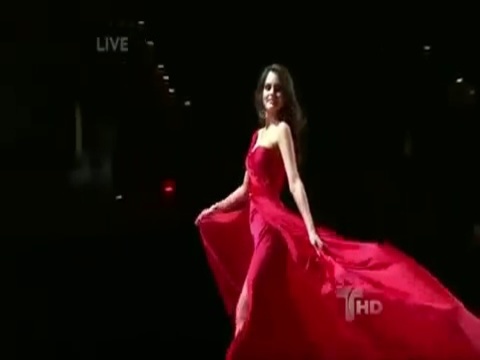}
   \includegraphics[width=0.32\columnwidth]{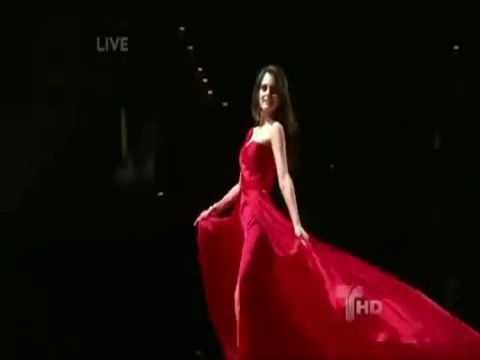} 
    \\\vspace{1mm}
    \includegraphics[width=0.32\columnwidth]{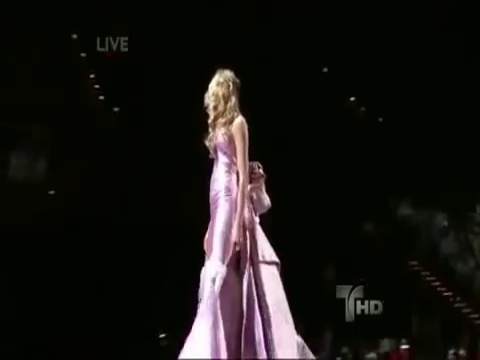}
    \includegraphics[width=0.32\columnwidth]{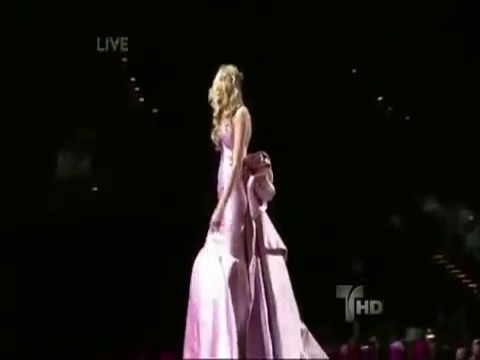}
    \includegraphics[width=0.32\columnwidth]{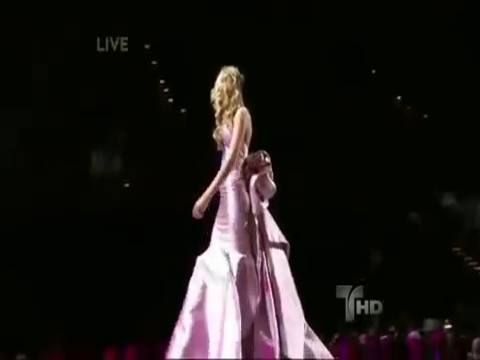} 
 \end{minipage}
\end{minipage}
 \caption{\small Examples of best and worst scores for Miss Universe versions 2003 and 2010}
\label{fig:cat_walk}
 \vspace{-2ex}
\end{figure*}

During the evening gown competition, candidates are given an average score based on their catwalk. 
Different judges are selected each year to score each candidate. 
This score is used in conjunction with the swimming competition, to select the best 5 finalists, 
where finally the Miss Universe winner is announced. 
Candidates with the best scores strut with attitude down the catwalk projecting confidence. 
See top row of Fig.~\ref{fig:cat_walk} for examples. 
Their arms are kept relaxed and swing naturally with the body. 
In general, they exhibit a flouncing walk and ooze elegance as they stalk the runway.
Candidates with the worst scores tend to exhibit issues such as stiff arms (resulting in robotic or awkward appearance)
and drooping their heads.
See bottom row of Fig.~\ref{fig:cat_walk} for examples. 
It can be also seen that the candidate 
with the worst catwalk during Miss Universe 2010 (Fig.~\ref{fig:cat_walk} bottom left) 
finds herself struggling to walk with the ribbon dress that is too tight for her.

Our central problem is to predict the best catwalk during 
the evening gown competition.
This can be considered as an instance of the ranking problem.
The ranking problem has been explored in various domains such as 
collaborative filtering, documents retrieval, and sentiment analysis~\cite{Cao2007}. 

In our work, we define two ranking sub-problems: (1) Miss Universe Listwise Ranking (MULR), and 
(2) Miss Universe Pairwise Ranking (MUPR).
While MULR focuses on rank ordering of all Miss Universe participants in the same year, MUPR considers pairwise comparisons of two participants in the same year.
We note that these two sub-problems have also been described in~\cite{Chapelle2009,WeiChen2009} for general machine learning problems.

\subsection{Miss Universe Listwise Ranking (MULR) Problem}

The MULR problem can be formalised as follows.
Given a query {\small $\mathcal{Q}_l = \{ \Mat{p}^{(q)}_{j}\}_{j=1}^{Nq}$}, where {\small $\Mat{p}^{(q)}_{j}$} is the video of a participant for Miss Universe from year {\small $q$} and {\small $N_q$} is the total number of candidates for that specific year. 
Let {\small $\mathcal{G}_l = \{ \mathcal{S}_m \}_{m=1}^{M}$} be the gallery containing {\small $M$} sets of Miss Universe from {\small $M$} years, 
where {\small $\mathcal{S}_m=\{ \Mat{p}_j^{(m)}\}_{j=1}^{N_m}$} is the set of {\small $N_m$} participant videos of Miss Universe from year {\small $m$}. 
Each set of participants {\small $\mathcal{S}_m$} is associated with a set of judgements (scores) {\small $\Vec{y}^{(m)} = [ y^{(m)}_1, \cdots, y^{(m)}_{N_m}]$}.
The judgement {\small $y^{(m)}_j$} represents the average score of participant {\small $\Mat{p}^{(m)}_j$}. We note that the average score is calculated by averaging the scores given by all the judges during the evening gown competition.
A set of video descriptors {\small $\Vec{v}^{(m)}_j = \Phi( \Mat{p}^{(m)}_j)$}, where {\small $\Vec{v}^{(m)}_j \in \mathbb{R}^d$} are extracted from each participant video, {\small $\Mat{p}^{(m)}_j$}.

Let {\small $f_l: \mathbb{R}^d \mapsto \mathbb{R}^1$} be a scoring function that calculates a participant score based 
on its corresponding video descriptors.
Given the query {\small $\mathcal{Q}_l$}, the function {\small $f_l$} can automatically score each participant in {\small $\mathcal{Q}_l$}.
Let {\small $\Vec{y}^{(q)} = [ y_1^{(m)}, \cdots, y_{N_q}^{(q)} ]$} be the actual score from the judges for participants in the query {\small $\mathcal{Q}_l$}, 
and {\small $\Vec{\hat{y}}^{(q)} = [ f(y)_1^{(m)}, \cdots,  f(y)_{N_q}^{(q)} ]$} be the estimated score of function {\small $f$} trained using the gallery set {\small $\mathcal{G}_l$}, the main task in MULR problem is to find the best {\small $f_l$}, where ideally the ranking of {\small $\Vec{\hat{y}}^{(q)}$} is the same as {\small $\Vec{y}^{(q)}$}.

\subsection{Miss Universe Pairwise Ranking (MUPR) Problem}

For the MUPR problem, we first consider a gallery, {\small $\mathcal{G}_p = \{ ( \Mat{p}^{(m)}_{l}, \Mat{p}^{(m)}_{k})\}^{M}_{m=1}, \Mat{p}^{(m)}_l, \Mat{p}^{(m)}_k \in \mathcal{S}_m, l \neq k$} 
wherein each element in the gallery is a pair of participant videos from the same year of Miss Universe.
Note that the gallery {\small $\mathcal{G}_p$} considered in this problem is different from 
the gallery {\small $\mathcal{G}_l$} considered in MULR problem.
Each pair in the gallery has its corresponding label {\small $y_{lk}^{(m)}$} which is defined via:

\begin{small}
\begin{equation}
y_{lk}^{(m)} = \left\{\begin{matrix}
 +1; &   y^{(m)}_{l} >  y^{(m)}_{k} \\ 
 -1, & \mbox{otherwise}   
\end{matrix}\right. ,
\label{eq:gt_mupr}
\end{equation}%
\end{small}%

\noindent
where {\small $y^{(m)}_{{l}}$} and  {\small $y^{(m)}_{k}$} are the actual score from the judges.
Let {\small $( \Mat{p}^{(q)}_{l}, \Mat{p}^{(q)}_{k})$, $y_{lk}^q$} be a query pair and its corresponding label, the main task for the MUPR problem is 
to find the best ranking function {\small $f_p ( \cdot ) = \{ -1 , +1 \}$} where ideally {\small $y_{lk}^{(q)} = f_p ( \Mat{p}^{(q)}_{l}, \Mat{p}^{(q)}_{k})$}.

\section{Proposed Approach}\label{sec:proposed_method}

We first describe the video descriptors used in our work.
We then present our approach to solve both MULR and MUPR problems simultaneously.

\subsection{Video Descriptors}\label{sec:descriptors}

Here, we describe how to extract from a video a set of features on a pixel level.
A video {\small ${\mathcal{V}} = \{ {\bm{I}}_t \}_{t=1}^T$} is an ordered set of {\small $T$} frames. Each frame {\small $\bm{I}_t \in \mathbb{R}^ {r\times c}$} can be represented by a set of feature vectors {\small $F_t = \{\bm{f}_p\}_{p=1}^{N_t}$}.
We extract the following $d=14$ dimensional feature vector for each pixel in a given frame~$t$~\cite{Joha2014b}:

\vspace{-2ex}
\begin{small}
\begin{equation}\label{eq:features}
\bm{f} = \left[ \; x, \; y, \; \bm{g}, \; \bm{o} \; \right]^\top
\vspace{-0.5ex}
\end{equation}%
\end{small}%

\noindent
where $x$ and $y$ are the pixel coordinates, while $\bm{g}$ and $\bm{o}$ are:
%
\begin{small}
\begin{eqnarray}
\hspace{-2ex} \bm{g} \hspace{-1ex} & = & \hspace{-1ex}
\left[ \; |J_x|, \; |J_y|, \; |J_{yy}|, \; |J_{xx}|, \; \sqrt{J_x^2 + J_y^2}, \; \text{atan} \frac{|J_y|}{|J_x|} \; \right]
\label{eq:features2}\\
\hspace{-2ex} \bm{o} \hspace{-1ex} & = & \hspace{-1ex}
\left[\; u, \;\;  v, \;\; \frac{\partial{u}}{\partial{t}}, \;\;
\frac{\partial{v}}{\partial{t}}, \;\;
\left (\frac{\partial u}{\partial x} + \frac{\partial v}{\partial y} \right ), \;\;
\left (\frac{\partial v}{\partial x} - \frac{\partial u}{\partial y} \right ) \;
\right]
\label{eq:features3}
\end{eqnarray}%
\end{small}%

The first four gradient-based features in Eq.~(\ref{eq:features2}) represent the first and second order intensity gradients at pixel location {\small $(x,y)$}.
The last two gradient features represent gradient magnitude and gradient orientation.
The optical flow based features in Eq.~(\ref{eq:features3}) represent:
the horizontal and vertical components of the flow vector,
the first order derivatives with respect to time,
the divergence and vorticity of optical flow~\cite{Ali2010}, respectively.
With this set of descriptors, we aim to capture the following attributes: 
shape with the coordinates, appearance with the gradients, and motion with the optical flow.
Typically only a subset of the pixels in a frame correspond to the object of interest ({\small $N_t < r \times c$}).
As such, we are only interested in pixels with a gradient magnitude greater than a threshold~{\small $\beta$}~\cite{KaiGuo2013}.
We discard feature vectors from locations with a small magnitude,
resulting in a variable number of feature vectors per frame.
For each video {\small ${\mathcal{V}}$}, the feature vectors are pooled into set {\small $\mathcal{F}=\{\bm{f}_n \}_{n=1}^N$} containing {\small $N$} vectors.

\subsection{Stacked Fisher Vectors}

The traditional Fisher Vector (FV) consists in describing a pooled set of features by its deviation from a generative model. 
FV encodes the deviations from a probabilistic version of a visual dictionary, which is typically a Gaussian Mixture Model (GMM)  with diagonal covariance matrices~\cite{Perronnin2010,JorgeSanchez2013}.
The parameters of a GMM with $K$ components can be expressed as {\small $\lambda=\{w_k,\bm{\mu}_k,\bm{\sigma}_k\}_{k=1}^{K}$},
where, {\small $w_k$} is the weight, {\small $\bm{\mu}_k$} is the mean vector, and {\small $\bm{\sigma}_k$} is the diagonal covariance matrix
for the $k$-th Gaussian.
The parameters are learned using the Expectation Maximisation algorithm~\cite{Bishop_PRML_2006} on training data.
Given the pooled set of features {\small $\mathcal{F}$} from video {\small ${\mathcal{V}}$}, the deviations from the GMM are then accumulated using~\cite{JorgeSanchez2013}:

\begin{small}
\begin{eqnarray}
\mathcal{G}_{\bm{\mu}_{k}}^{\mathcal{F}}    & = & \frac{1}{N\sqrt{w_k}} \sum\nolimits_{n=1}^{N} \gamma_n(k)\left( \frac{\bm{f}_n - \bm{\mu}_k}{\bm{\sigma}_k} \right)\\
\mathcal{G}_{\bm{\sigma}_{k}}^{\mathcal{F}} & = & \frac{1}{N\sqrt{2w_k}} \sum\nolimits_{n=1}^{N} \gamma_n(k)\left[ \frac{\left(\bm{f}_n - \bm{\mu_k}\right)^2}{\bm{\sigma}_k^2} -1 \right]
\end{eqnarray}%
\end{small}%

\noindent where vector division indicates element-wise division
and {\small $\gamma_n(k)$} is the posterior probability of {\small $\bm{f}_n$} for the $k$-th component:

\vspace{-1ex}
\noindent
\begin{small}
\begin{equation}
\gamma_n(k) = \frac{w_k\mathcal{N}(\bm{f}_n|\bm{\mu}_k, \bm{\sigma}_k)}{\sum\nolimits_{i=1}^{K}w_i\mathcal{N}(\bm{f}_n|\bm{\mu}_i, \bm{\sigma}_i)}
\end{equation}%
\end{small}%

\noindent
The Fisher vector for each video ${\mathcal{V}}$ is represented as the concatenation of
{\small $\mathcal{G}_{\bm{\mu}_{k}}^{\mathcal{F}}$} and {\small $\mathcal{G}_{\bm{\sigma}_{k}}^{\mathcal{F}}$} (for {\small $k$~=~$1, \ldots, K$}) into vector~$\mathcal{G}_{\lambda}^{\mathcal{F}}$.
As {\small $\mathcal{G}_{\bm{\mu}_{k}}^{\mathcal{F}}$} and {\small $\mathcal{G}_{\bm{\sigma}_{k}}^{\mathcal{F}}$} are \mbox{$d$-dimensional},
{\small $\mathcal{G}_{\lambda}^{\mathcal{F}}$} has the dimensionality of {\small $2dK$}.
Power normalisation is then applied to each dimension in {\small $\mathcal{G}_{\lambda}^{\mathcal{F}}$}.
The power normalisation to improve the FV for classification was proposed in~\cite{Perronnin2010} of the form {\small $z \leftarrow \text{sign}(z)|z|^\rho$},  where {\small $z$} corresponds to each dimension and the power coefficient {\small $\rho =1/2$}. Finally, {\small $l_2$}-normalisation is applied. 
Note that we have omitted the deviations for the weights as they add little information~\cite{JorgeSanchez2013}.

Stacked Fisher Vectors (SFV) is a multi-layer representation of standard FV~\cite{Peng2014eccv}. 
SFV first performs traditional FV representation over densely sampled subvolumes based on low level descriptors. 
The extracted FVs have a high dimensionality and are fed the next layer. The second layer reduces the obtained FVs, and then those reduced FVs are encoded again with FV representation.

\subsection{Classification}\label{sec:classification}

We address both MULR and MUPR problems using the same framework. 
Recall that the main objective of the MULR problem is to find the best {\small $f_l(\cdot)$} wherein its scores can be used to rank 
the Miss Universe participants from the same year.
We model such a function as a linear regression function:

\begin{small}
\begin{equation}
f_l(\Vec{v}) = \Vec{w}^{\top}\Vec{v} + b , 
\label{eq:fl_model}
\end{equation}%
\end{small}%

\noindent
where {\small $\Vec{w}$} and {\small $b$} are the parameters of the regression model and {\small $\Vec{v} \in \mathbb{R}^d$} is the extracted video descriptor after applying SFV.
As it is not trivial to train the regression given the gallery $\mathcal{G}_l$ with its corresponding actual ranking, we solve this 
problem by addressing MUPR, which is a much easier problem.
This is possible as the ranking function $f_p$ can be defined in terms of the scoring function $f_l$:

\begin{small}
\begin{equation}
	f_p(\Vec{v}_l, \Vec{v}_k) = \operatorname{sign} ( f_l ( \Vec{v}_l) - f_l (\Vec{v}_k ) ) ,
\end{equation}%
\end{small}%

\noindent
where {\small $\operatorname{sign} ( \cdot )$} only takes the sign of the input.
Plugging the scoring function model into the above equation we obtain:

\noindent
\begin{small}
\begin{align}
    f_p(\Vec{v}_l, \Vec{v}_k) = & \operatorname{sign} ( f_l ( \Vec{v}_l) - f_l (\Vec{v}_k ) ) \nonumber \\
	 & \operatorname{sign}( \Vec{w}^{\top} \Vec{v}_l + b - \Vec{w}^{\top} \Vec{v}_k - b ) \nonumber \\
	 & \operatorname{sign}( \Vec{w}^{\top} (\Vec{v}_l - \Vec{v}_k)  ) \nonumber \\
	 & \operatorname{sign}( \Vec{w}^{\top} \Vec{z}  ) , 
\end{align}%
\end{small}%

\noindent where {\small $\Vec{z} \in \mathbb{R}^d$} is the new descriptor extracted via: {\small $\Vec{z} = \Vec{v}_l - \Vec{v}_k$}.
Notice that both {\small $f_l$} and {\small $f_p$} share the same model parameter~{\small $\Vec{w}$}. 
As we only focus on the ranking for MULR problem, the bias parameter, {\small $b$} in Eq.~(\ref{eq:fl_model}) can be excluded; the regression model thus becomes:

\begin{small}
\vspace{-1ex}
\begin{equation}
f_l(\Vec{v}) = \Vec{w}^{\top}\Vec{v} .
\end{equation}%
\end{small}%

With the above modification, we only need to perform the training step once for both functions. 
To this end, we perform the training step for the ranking function,~{\small $f_p$}. 
Following the training formulation from the RankSVM described in~\cite{Joachims2006}:

\begin{small}
\begin{equation}
	\frac{1}{2} \| \Vec{w} \|_2^2 + C \sum\nolimits_{m=1}^M \sum\nolimits_{\Vec{v}^{(m)}_l, \Vec{v}^{(m)}_k \in \mathcal{S}_m} \ell (y^{(m)}_{lk} \Vec{w}^{\top} \Vec{z}^{(m)}_{lk} ) , 
\end{equation}%
\end{small}%

\noindent where {\small $\Vec{z}^{(m)}_{lk} = \Vec{v}^{(m)}_{l} - \Vec{v}^{(m)}_{k}$}, is the new descriptor as described above; $y^{(m)}_{lk}$ is the ground truth for the MUPR problem described in Eq.~(\ref{eq:gt_mupr}),  {\small $C$} is a training parameter and {\small $\ell ( \cdot )$} is the hinge loss.

\section{Miss Universe (MU) Dataset}\label{sec:dataset}

\begin{figure}[!b]
\centering 
{\includegraphics[width=0.4\columnwidth]{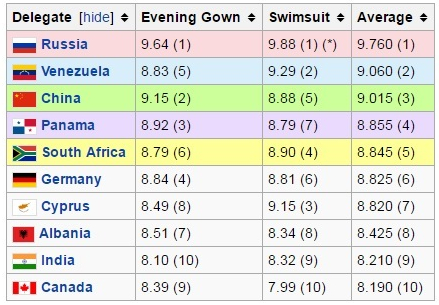}}
~
{\includegraphics[width=0.4\columnwidth]{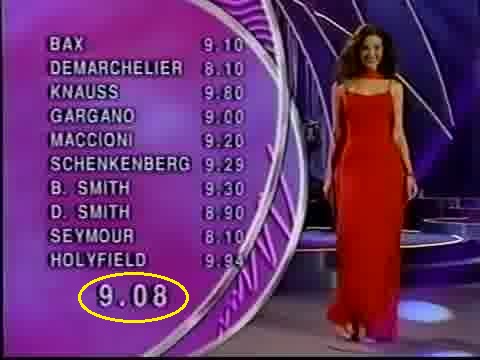}}
\caption{\small Judges' scores. Left: from Wikipedia. Right: from the video.}
 \label{fig:judges_scores}
 \vspace{-3.5ex}
\end{figure}

In this work, we propose the Miss Universe Dataset to address our problems.
In particular, we have collected a novel dataset of videos depicting the evening 
gown competition for 10 years of Miss Universe (MU). 
The videos span from 1996 to 2010, where the judges' scores are available. 
The videos were downloaded from YouTube and the scores were obtained from the videos themselves or Wikipedia. 
Fig.~\ref{fig:judges_scores} shows examples of scores. 
While the scores taken from the videos include each individual score from judge, only the average is used (circled in yellow).

We have collected 105 videos, 18,343 frames in total, with an average of 175 per video. Each video shows a candidate during the evening gown competition. Additionally, we manually select the bounding box enclosing each participant.

\begin{figure}[!b]
\centering
\stackunder[2pt] 
 {\includegraphics[width=0.19\columnwidth]{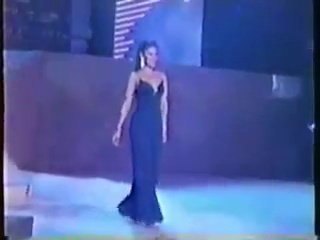}}
{\footnotesize 1996}
\hspace{-1.5ex}
\stackunder[2pt] 
 {
\includegraphics[width=0.19\columnwidth]{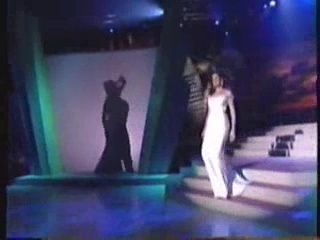}}
{\footnotesize 1997}
\hspace{-1.5ex}
\stackunder[2pt] 
 {
\includegraphics[width=0.19\columnwidth]{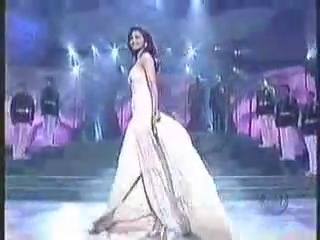}}
{\footnotesize 1998}
\hspace{-1.5ex}
\stackunder[2pt] 
 {
 \includegraphics[width=0.19\columnwidth]{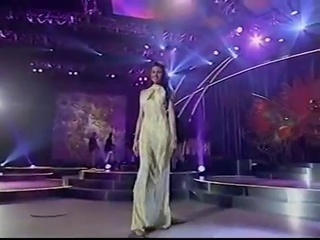}}
{\footnotesize 1999}
\hspace{-1.5ex}
\stackunder[2pt] 
 {
\includegraphics[width=0.19\columnwidth]{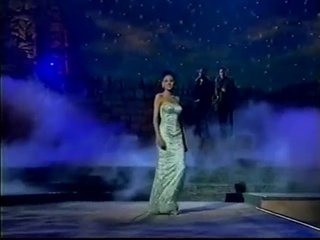}}
{\footnotesize 2000}
\\
\vspace{1.5ex}
\stackunder[2pt] 
 {\includegraphics[width=0.19\columnwidth]{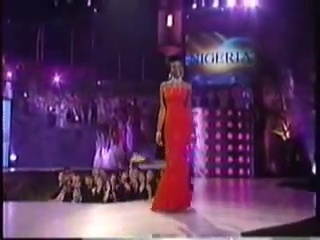}}
{\footnotesize 2001}
\hspace{-1.5ex}
\stackunder[2pt] 
 {
\includegraphics[width=0.19\columnwidth]{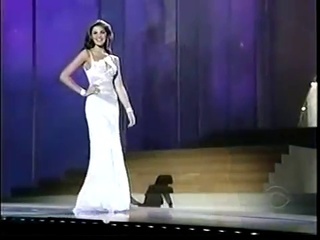}}
{\footnotesize 2000}
\hspace{-1.5ex}
\stackunder[2pt] 
 {
\includegraphics[width=0.19\columnwidth]{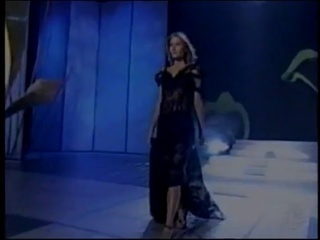}}
{\footnotesize 2003}
\hspace{-1.5ex}
\stackunder[2pt] 
 {
 \includegraphics[width=0.19\columnwidth]{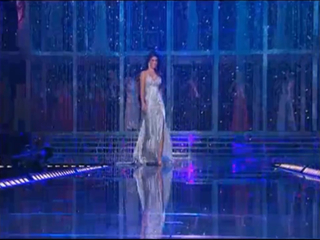}}
{\footnotesize 2007}
\hspace{-1.5ex}
\stackunder[2pt] 
 {
\includegraphics[width=0.19\columnwidth]{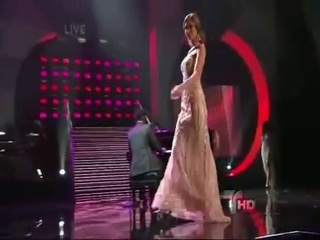}}
{\footnotesize 2010}
\caption{\small Catwalk stages for all years}
\label{fig:stage}
\vspace{-1ex}
\end{figure}

It is noteworthy to mention that the proposed MU dataset is extremely challenging due to variations in capture conditions  for each year: (1) catwalk stage; (2) illumination conditions; (3) cameras capturing the event. 
As for the variations in cameras capturing the event, for our purpose we opted to use only one camera view depicting the longest walk without interruptions. Fig.~\ref{fig:stage} shows the catwalk stage for each year in the MU dataset.
The dataset is available from~{\href{http://www.itee.uq.edu.au/sas/datasets}{\small\bf\texttt{http://www.itee.uq.edu.au/sas/datasets}}}

\subsection{Evaluation Protocol}\label{sec:protocol}

We use leave-one-year-out protocol as the evaluation protocol for both MUPR and MULR.
In particular, for each training-test set, we consider all participants from one year as the testing and the rest as training. 
As the dataset covers ten years of Miss Universe videos, there are ten training-test sets.
Once the results from all the ten training-test sets are determined, the performance of a method is reported as the average of these results.

\textbf{The MULR problem evaluation metric --- } In the MULR problem we are interested in evaluating how similar is the ranking determined to the scoring function $f_l$ from the actual ranking of each year.
To this end, we use the Normalized Discount Cumulative Gain (NDCG) proposed to measure ranking quality of documents~\cite{CPLee2014,Qin2010}.
NDCG is often used to measure of the efficacy of  web search algorithms~\cite{CPLee2014}.
To use this metric, we consider each candidate video as a ``visual'' document. 
Here the rating of each visual document corresponds to the rank of the participant. 
Thus, we rate each visual document/participant video by assigning values between 1 to 10 with 10 being the highest score and 1 for the lowest. 
These values are assigned according to their corresponding rank.
For instance, we assign the participant having the highest score with value 10 and assign the runner up with value 9.

In the original formulation, the NDCG measures the ranking quality based on the top $b$ rated documents~\cite{CPLee2014}:

\begin{small}
\begin{equation}
\text{NDCG}_{@b} =  {\text{DCG}_{@b}} / {\text{IDCG}_{@b}},
\label{eq:listwise}
\end{equation}%
\end{small}%

\noindent where \text{DCG} is the discounted cumulative gain at particular rank position $b$ and is defined as:

\begin{small}
\begin{equation}
\text{DCG}_{@b} =  \sum\nolimits_{j=1}^{b} \frac{ (2^{r(j)} - 1) }{\log_2 ( \max(2,j)  )}
\end{equation}%
\end{small}%

The rating of the {\small $j$}-th participant in the ranking list is given by {\small $r(j)$} and {\small $\text{IDCG}_{@b}$} is the ideal DCG at position $b$.  Note that {\small $b = 1,\cdots, N_q$} with {\small $N_q$} being the length of the ordering. 
A perfect list gets a {\small $\text{NDCG}_{@b}$} score of {\small $1$}.
For our case, we always set {\small $b=N_q$}.
We report the average percentage {\small $\text{NDCG}_{@Nq}$} over all partitions and refer to it as the {\small $\text{NDCG}$}.

\textbf{The MUPR problem evaluation metric --- } For the MUPR problem, we use the modified Kendall's {\small $\mathcal{K}_\tau$} as a performance measure discussed in~\cite{Joachims2002}. 
{\small $\mathcal{K}_{\tau}$} is defined as the number $C$ of concordant pairs and the number $D$ of discordant pairs. A pair {\small $(p_l^{(m)}, p_k^{(m)})$} with {\small $l \neq k$} is concordant, if  {\small $\hat{y}^{(m)}_{lk} = y^{(m)}_{lk}$}. It is discordant if they disagree. The sum of $C$ and $D$ must be $\binom{N_q}{2}$. Kendall's $K_\tau$ can be defined as:

\begin{small}
\begin{equation}
\mathcal{K}_{\tau} = \frac{C - D}{C + D} = 1- \frac{2D}{\binom{N_q}{2}}
\label{eq:pairwise}
\end{equation}%
\end{small}%
\vspace{-1ex}
\section{Experiments}
\label{sec:experiments}

To the best of our knowledge, this is the first work to study catwalk analysis for Miss Universe.
We used the new Miss Universe dataset containing 10 versions of Miss Universe.
Miss Universe 2003 contains 15 participants. The remaining versions each contain 10 participants.
We used the bounding box enclosing the participant provided with the dataset.
We resized all bounding boxes to $100\times 50$.

\textbf{Setup --- } All videos were converted into gray-scale.
We use the leave-one-year-out protocol, where we leave one version of Miss Universe out for testing.
For each video, we extract a set of {\small $d=14$} dimensional features as explained in Section~\ref{sec:descriptors}.  Based on~\cite{Joha2014b}, we used {\small $\beta=40$}, where {\small $\beta$} is the threshold  used for selecting interesting low-level features.
Parameters for the visual vocabulary GMM  were learned using a large set of descriptors randomly obtained from training videos using the iterative Expectation-Maximisation algorithm~\cite{Bishop_PRML_2006}.
The systems were implemented with the aid of the Armadillo C++ library~\cite{Armadillo2016}.
Experiments were performed with three separate GMMs with varying number of components {\small $K = \{256,512,1024\}$}.

For the traditional FV representation, each video is represented by a FV. 
The FVs are fed to a linear SVM for classification.

For the first layer of SFV, we obtained a varying number of vectors using the traditional FV representation. Each vector is obtained using the low-level descriptors of $5$ consecutive frames. Then, we advanced by a frame and obtained a new FV.
For the second layer of SFV, we reduced the dimensionallity of each vector from layer 1 using two methods: Principal Component Analysis (PCA)  and Random Projection (RP). For PCA, we retained the $90\%$ of the energy~\cite{Amatriain2011}. For RP, we used the resulting dimensionallity number obtained by PCA. We referred to these methods as SFV-PCA and SFV-RP.

Our classification model is described in Section~\ref{sec:classification}. As explained, we address both problems using the same framework. 
We solve MULR by addressing MUPR first. 
In our implementation, we solve MUPR by using the LibLinear package~\cite{FanRE2008} and set the bias parameter {\small $b$} to 0.

\textbf{Results for MUPR --- } In Table~\ref{tab:MUPR}, we present the results for MUPR. 
The evaluation metric employed is Kendall's $\mathcal{K}_{\tau}$ as per Eq.~(\ref{eq:pairwise}). 
From this table, we can see that our classification models using both dimensionallity reduction techniques outperform the baseline FV representation. 
Using SFV-PCA with a visual dictionary size of $256$ Gaussians leads to the best performance of $56.73\%$, which is $3.05$ points higher than SFV-RP. 
PCA is an essential step for dimensionality reduction for this application. Despite the simplicity of random projection, its performance is inferior to PCA.

\begin{table}[!tb]
\caption{Results for MUPR using  $\mathcal{K}_\tau$} 
\centering 
\small
\begin{tabular}{l ccc} 
\hline \hline 
 \multirow{ 2}{*}{Method}  &\multicolumn{3}{c}{Visual Vocabulary Size} \\
 \cline{2-4}
   & 256 & 512 & 1024 \\ 
\hline
  FV (baseline) &  $52.63~\%$ 		& $52.16~\%$ & $52.03~\%$ \\
  SFV-PCA  		&  $\bm{56.73}~\%$ 	& $51.81~\%$ & $50.98~\%$ \\
  SFV-RP   		&  $53.68~\%$ 		& $46.92~\%$ & $47.87~\%$ \\
\hline 
\end{tabular}
\label{tab:MUPR}
\vspace{-1ex}
\end{table}

\begin{table*}[!tb]
\caption{NDCG for each year using best settings for SFV-PCA}
\centering 
\small
\begin{tabular}{lcccccccccc} 
\hline
{\bf Year}  & 2010 & 2007 & 2003 & 2002  & 2001  & 2000  & 1999 & 1998 & 1997 & 1996\\
{\bf NDCG} & $\bm{77.71}~\%$  & $\bm{78.95}~\%$   &	 $\bm{52.97}~\%$   &	 $\bm{62.78}~\%$    &	 $44.37~\%$  &	 $44.71~\%$  &	 $\bm{82.93}~\%$  &	 $\bm{87.02}~\%$  &	 $51.69~\%$  &	 $\bm{77.38}~\%$ \\
\hline 
\end{tabular}
\label{tab:MULR_run}
\vspace{-1ex}
\end{table*}

\begin{figure}
\centering
{\includegraphics[width=0.5\columnwidth]{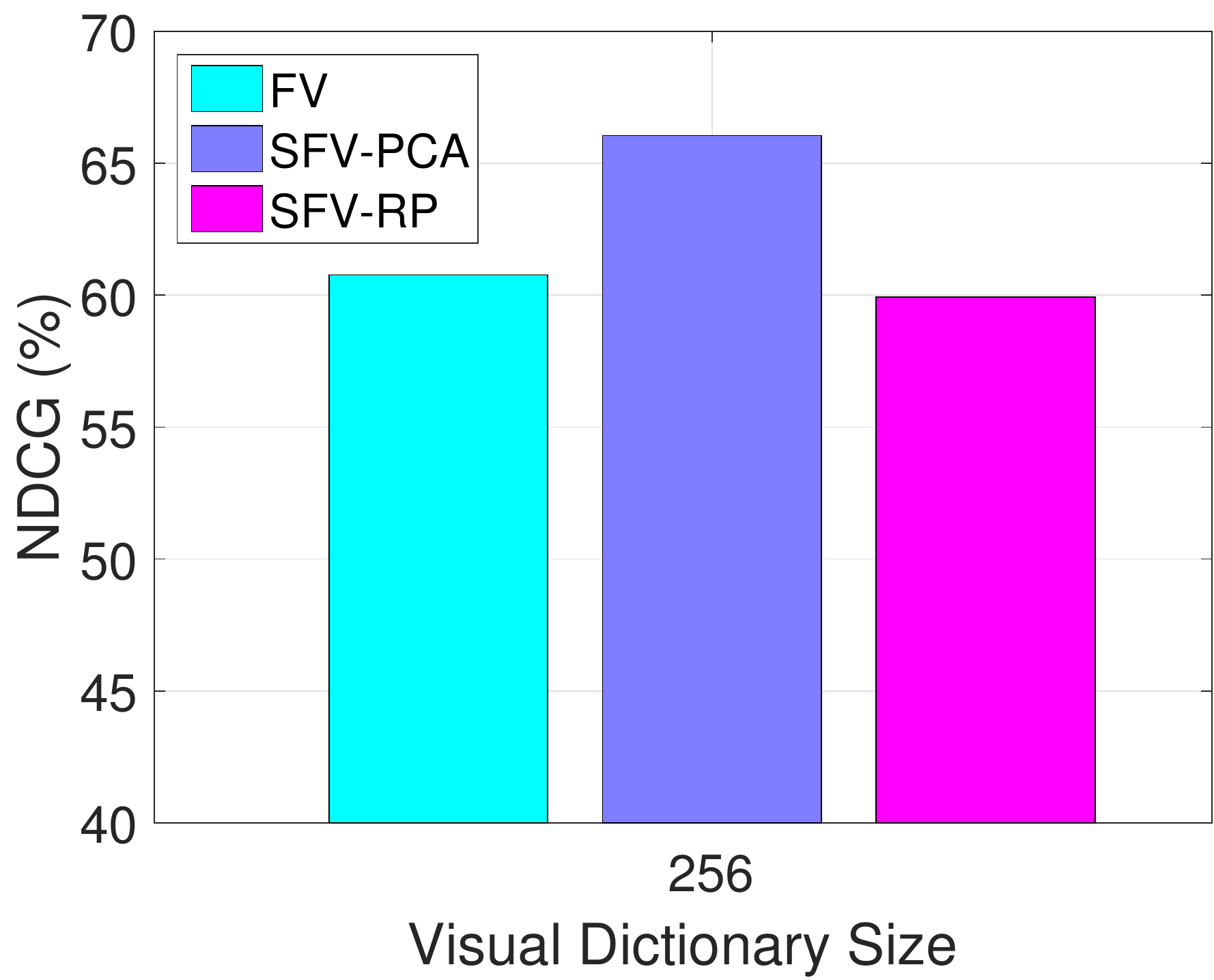}}
\caption{\small Results for MULR using NDCG}
 \label{fig:scores}
  \vspace{-2ex}
\end{figure}

\textbf{Results for MULR --- } Using the best setting for MUPR obtained with a visual vocabulary size of $256$ Gaussians, we evaluated MULR.
The evaluation metric employed is NDCG as per Eq.~(\ref{eq:listwise}).  
Fig.~\ref{fig:scores} shows that SFV-PCA attained the best performance with $66.05\%$.
Table~\ref{tab:MULR_run} shows the individual performance using NDCG for each of the ten training-test sets as explained.
Our SFV-PCA classification approach shows a performance which is higher than $50\%$ in 7 out 10 training/test sets. In 2 out of the 10  training/test sets we obtained a performance higher than $82\%$.
Moreover, our Miss Universe automatic prediction system was able to recognise the winner for the evening gown competition for years 1998 and 1999, which explains the higher performance for those years as in NDCG  top ranked instances are considered more important. The predicted winner is also found in the top 3 for 5 out 10 versions of Miss Universe (2010, 2007, 1999, 1998, and 1996).

\section{Main Findings and Future Work}\label{sec:conl}

In this work, we have present a promising approach to automatically detect the winner during the evening gown competition of Miss Universe. 
To this end, we have created a new dataset comprising 10 years of the evening gown competition selected from 1996 to 2010.
We addressed this problem using action analysis techniques. 
We defined two problems that are of potential interest for the beauty pageant industry and the fashion industry. 
In the former problem, we are interested in predicting the winner of the competition, which can be also of interest for specialised betting sites.
The fashion industry can have an innovative automatic system to compare two catwalks that can be used as training system for amateur models.
Our system for predicting the winner of the evening gown competition shows we are able to rank the winner in the top 3 best predicted scores in $50\%$ of the cases.

For future work, we propose to enlarge the dataset, to extend to the swimsuit catwalk competition, and the use of pose features. 
The current dataset can be enlarged using other Miss Universe versions, other beauty pageant competitions, and catwalks from international fashion trade shows. Given that scores are not always publicly available, an online competitive Catwalk rating game can be designed similar to the style rating game called \textit{Hipster Wars}~\cite{TamaraBergEccv2014}. With this online game it would be possible to crowd source reliable human judgements of catwalks.
The swimsuit catwalk competition together with the evening gown competition are critical to the selection of  the next Miss Universe. For the swimming competition, other attributes apart from the catwalk would be needed to take into consideration such as good muscle tone, body proportion, body fat, body shape, and fitness. All those attributes are also visual attributes.
Pose is an important attribute for catwalks. 
We envisage that pose-based Convolutional Neural Network features in conjunction with IDT can increase our system performance.
This combination has been recently shown to be effective for action recognition~\cite{Cheron2015}.

Finally, we note that this work can be extended to other applications that require action assessment. For instance, patient rehabilitation and high performance sports. In both cases, an automatic system able to evaluate the progress of a patient or an athlete would be valuable.

\balance
\footnotesize
\bibliographystyle{abbrv}
\bibliography{references}
\end{document}